\documentclass{article}

\usepackage{microtype}
\usepackage{graphicx}
\usepackage{amsmath} 
\usepackage{amssymb}
\usepackage{multirow}
\usepackage{booktabs} 
\usepackage{caption}
\usepackage{subcaption}
\usepackage{dcolumn}
\newcolumntype{d}[1]{D{.}{.}{#1}}

\usepackage{hyperref}
\usepackage{siunitx}

\usepackage{algcompatible}
\usepackage[compatible]{algpseudocode}

\usepackage[accepted]{mlsys2023}

\mlsystitlerunning{Arithmetic Intensity Balancing Convolution for Hardware-aware Efficient Block Design}

\begin{document}

\twocolumn[
\mlsystitle{Arithmetic Intensity Balancing Convolution for Hardware-aware Efficient Block Design}




\begin{mlsysauthorlist}
\mlsysauthor{Shinkook Choi}{nota}
\mlsysauthor{Junkyeong Choi}{nota}
\end{mlsysauthorlist}

\mlsysaffiliation{nota}{Nota Inc. Seoul, Republic of Korea}

\mlsyscorrespondingauthor{Shinkook Choi}{shinkook.choi@nota.ai}

\mlsyskeywords{Machine Learning, MLSys}

\vskip 0.3in

\begin{abstract}
As deep learning advances, edge devices and lightweight neural networks are becoming more important. To reduce latency in the AI accelerator, it's essential to not only reduce FLOPs but also enhance hardware performance. We proposed an arithmetic intensity balancing convolution (ABConv) to address the issue of the overall intensity being limited by the small weight arithmetic intensity for convolution with a small spatial size. ABConv increased the maximum bound of overall arithmetic intensity and significantly reduced latency, without sacrificing accuracy. We tested the latency and hardware performance of ABConv on the Arm Ethos-U65 NPU in various configurations and used it to replace some of MobileNetV1 and ResNet50 in image classification for CIFAR100.
\end{abstract}
]



\printAffiliationsAndNotice{}  

\section{Introduction}
\label{Introduction}
Over the last few years, the adoption of Artificial Intelligence (AI) has grown significantly, resulting in a huge advancement in AI. However, as AI capabilities have increased, so has the complexity of algorithm. 
Since most high-performance AI models are too large for edge devices to manage, designing AI model optimization technology has become more important than ever to deal with stringent memory and power constraints.

Many studies are reported that the lightweight deep neural networks (DNN) are designed to reduce the number of float-point operations (FLOPs) and the number of parameters for executing on edge devices \cite{howard2017mobilenets,sandler2018mobilenetv2,tan2019efficientnet,tan2021efficientnetv2}. However, the indirect metrics of computation complexity such as FLOPs and parameters are not directly correlated with how efficiently the hardware can compute on edge devices \cite{akin2022searching, ma2018shufflenet}.
Thus it is important to consider both the characteristics of the neural network and the target hardware when designing an optimized neural network

To design the most efficient deep learning model for the hardware, the concept of arithmetic intensity \cite{harris2005mapping} and roofline analysis \cite{williams2009roofline} is applicable to formulate data reuse and hardware performance.
Arithmetic intensity is the ratio of the number of arithmetic operations to the memory footprint. 
Higher arithmetic intensity implies the degree of data reuse, resulting in efficient computation.
As DNN workloads emerge as a primary concern in the sense of computation efficiency \cite{jouppi2017datacenter,park2018deep,yang2017method} utilizes roofline analysis using arithmetic intensity in order to visualize the efficiency of DNN workloads.
Furthermore, Jha et al.\cite{jha2020modeling} explored more on the concept of arithmetic intensity and data reuse, splitting data reuse into weight reuse and activation reuse according to the types of data used for DNN.

Another hardware characteristic we considered while designing the network is staircase pattern of the convolution latency\cite{tang2021bridge}. 
Even though the arithmetic complexity of convolution increases linearly, the latency of the convolution does not linearly scale.
It rather scales up with a certain step size, which varies from hardware to hardware, showing staircase latency pattern. 

\begin{table}[b!]
\centering
\caption{Data reuse characteristics of the standard convolution, the group convolution, and the proposed ABConv and ABConv-exp.}
\label{table:arithmetic_intensity}
\resizebox{\linewidth}{!}{%
\begin{tabular}{c|c|c|c|c|c} 
\toprule
\multirow{2}{*}{} & \multirow{3}{*}{\begin{tabular}[c]{@{}c@{}}MACs\\$(M_c)$\end{tabular}} & \multirow{3}{*}{\begin{tabular}[c]{@{}c@{}}Weight size\\$(W)$\end{tabular}} & \multirow{3}{*}{\begin{tabular}[c]{@{}c@{}}Activation size\\$(A)$\end{tabular}} & \multicolumn{2}{c}{Arithmetic intensity}                                                                                   \\ 
\cline{5-6}
                  &                                                                        &                                                                             &                                                                                 & \begin{tabular}[c]{@{}c@{}}Weight\\$(M_c/W)$\end{tabular} & \begin{tabular}[c]{@{}c@{}}Activation\\$(M_c/A)$\end{tabular}  \\ 
\toprule
Standard conv     & $S_o^2k^2 C_{in} C_{out}$                                              & $k^2 C_{in} C_{out}$                                                        & $S_o^2(C_{in}+C_{out} )$                                                        & $S_o^2$                                                   & $\dfrac{k^2C_{in} C_{out}}{C_{in}+C_{out} }$                  \\ 
\hline
Group conv        & $\dfrac{S_o^2k^2 C_{in} C_{out}}{g}$                                   & $\dfrac{k^2 C_{in} C_{out}}{g}$                                             & $S_o^2(C_{in}+C_{out} )$                                             & $S_o^2$                                                   & $\dfrac{k^2C_{in} C_{out}}{g(C_{in}+C_{out}) }$                \\ 
\hline
ABConv            & $\dfrac{S_o^2k^2 C_{in} C_{out}}{g}$                                   & $\dfrac{k^2 C_{in} C_{out}}{g^2}$                                           & $S_o^2(C_{in}+C_{out} )$                                                        & $gS_o^2$                                                  & $\dfrac{k^2C_{in} C_{out}}{g(C_{in}+C_{out})}$                 \\ 
\hline
ABConv-exp        & $S_o^2k^2 C_{in} C_{out}$                                              & $\dfrac{k^2 C_{in} C_{out}}{g}$                                             & $S_o^2(C_{in}+2gC_{mid}+C_{out} )$                                              & $gS_o^2$                                                  & $\dfrac{k^2C_{in} C_{out}}{C_{in}+2gC_{mid}+C_{out} }$         \\
\bottomrule
\end{tabular}
}
\end{table}

In this study, we highlight a problem that occurs on Arm Ethos-U65 NPU when the arithmetic intensity is bounded to the weight data due to the input spatial size of the convolution layer. To solve the problem, we propose a new layer that balances the arithmetic intensity of weight and activation data, thereby increasing the overall arithmetic intensity.
We demonstrate that increasing the weight arithmetic intensity improves the hardware performance and evaluate the effectiveness of our method in reducing inference speed. Our method also includes a hardware-adaptive, staircase-aware method for fine-tuning the model for Ethos-U65 NPU.
Our experiments show that the model using the proposed layer achieves better accuracy and lower latency than both MobileNetV1 and ResNet50 for image classification tasks.
To the best of our knowledge, this is the first work that shows that the hardware performance of the convolution operation can lack a certain type of arithmetic intensity. 

\section{Motivation}
In deep neural networks, arithmetic intensity is the number of multiply-accumulate operations (MACs) performed on each byte of weights (filter coefficients) and activations (input feature maps and output feature maps). 
When MACs, weight size, activation size is expressed $M_c$, $W$ and $A$, respectively, the overall arithmetic intensity ($\frac{M_c}{W+A}$) has upper bounds on both weight arithmetic intensity ($\frac{M_c}{W}$) and activation arithmetic intensity ($\frac{M_c}{A}$), as shown in Equation \ref{Eq:arithmetic_intensity_lowerbound}.
Therefore, if any one of weight arithmetic intensity and activation arithmetic intensity becomes smaller, the upper limit of whole arithmetic intensity decreases, and thus the whole arithmetic intensity is reduced. 

\begin{equation}
\label{Eq:arithmetic_intensity_lowerbound}
\frac{M_c}{W+A} < \frac{M_c}{W}, \frac{M_c}{W+A} < \frac{M_c}{A} \quad \text{if } M_c,W,A>0.
\end{equation}

Table~\ref{table:arithmetic_intensity} shows data reuse characteristics of convolution. The weight arithmetic intensity is determined by spatial size $(S_o)$ and the activation arithmetic intensity is calculated by kernel size ($k$), input channel size $(C_{in})$, and output channel size $(C_{out})$.

Many neural networks based on convolution layers are designed to reduce the spatial size and increase the channel size as the network depth increases. This design results in a decrease in overall arithmetic intensity, which is mainly bounded to weight arithmetic intensity in the later stages of the network. Therefore, it is necessary to design a new layer expanding the spatial size and increasing data reuse of weight to fully utilize the hardware.

\section{Proposed method}
\begin{figure}[t]
\centering
\begin{subfigure}[b]{1\columnwidth}
         \centering
         \includegraphics[width=0.8\columnwidth]{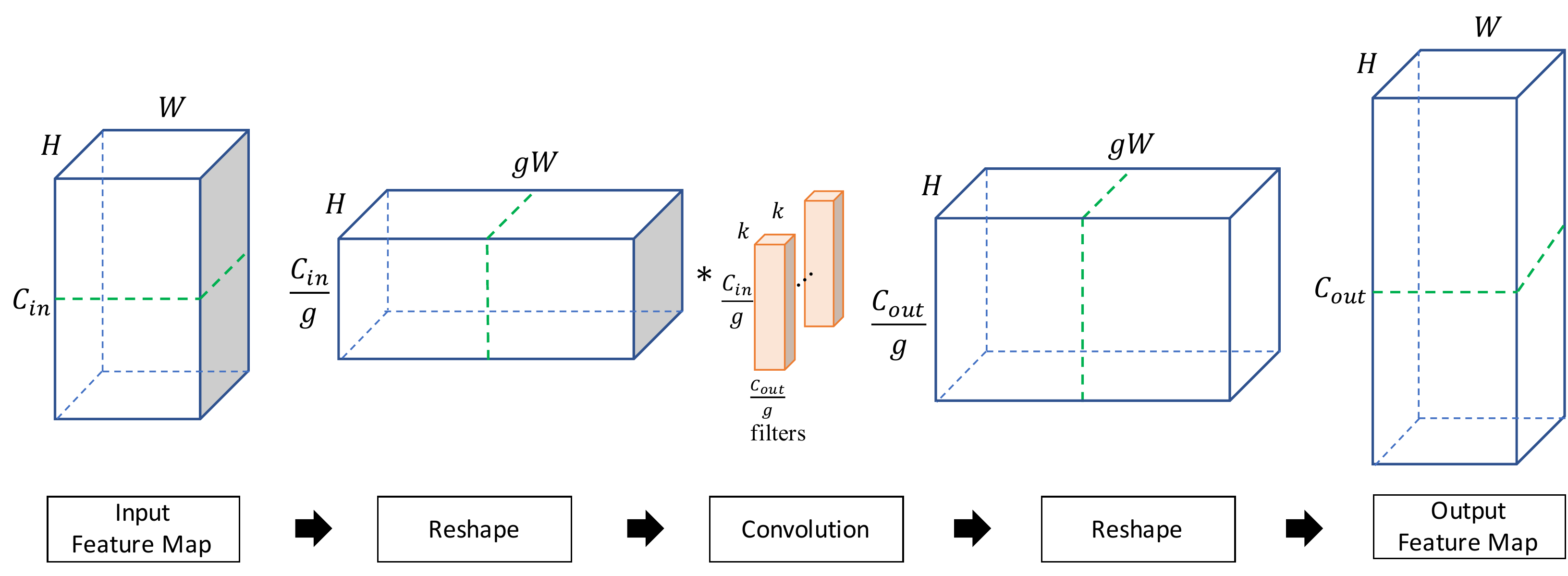}
         \caption{ABConv}
 \end{subfigure}
\begin{subfigure}[b]{1\columnwidth}
         \centering
		\includegraphics[width=1\columnwidth]{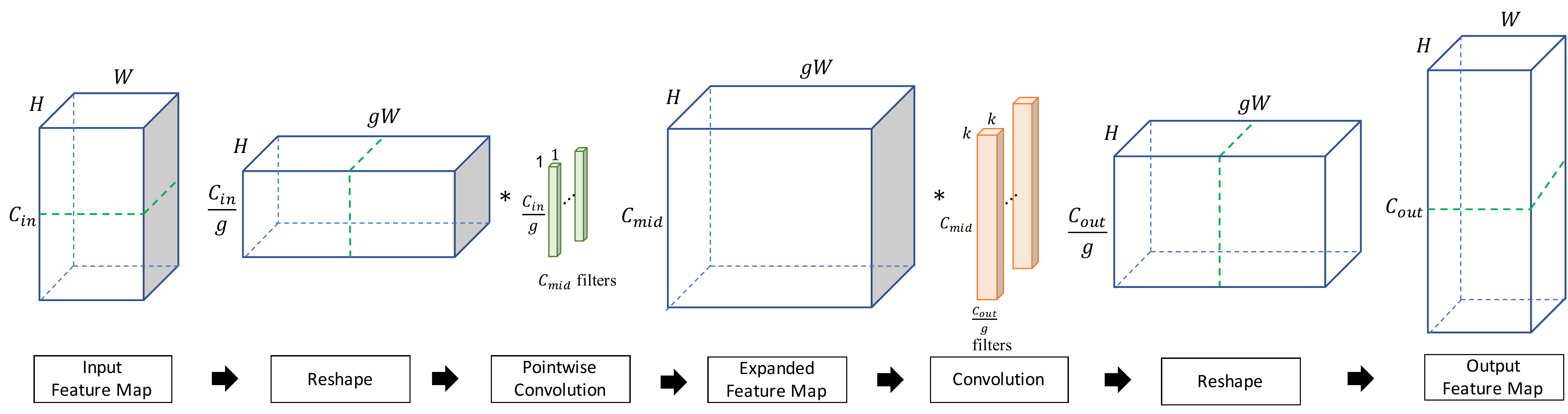}
         \caption{ABConv-exp}
     \end{subfigure}
\caption{An illustration of the proposed ABConv and ABConv-exp for outputting the same number of feature maps as the standard convolution layer.}
\label{Fig:ABConv}
\end{figure}

    \subsection{Arithmetic intensity Balancing Convolution}
    We propose arithmetic intensity balancing convolution (ABConv) to maximize hardware utilization. When the whole arithmetic intensity is bounded by the weight arithmetic intensity, ABConv expands the spatial size and reuses more weight data to increase the whole arithmetic intensity.
    As shown in Figure~\ref{Fig:ABConv}, ABConv consists of three layers which are a reshape layer, a convolution layer, and a reshape layer in order. The first reshape layer cuts the feature map by groups $(g)$ in the channel direction and concatenates them in the spatial size direction. In the convolution layer, the convolution kernel which is divided by groups is convolved with the input feature map whose spatial size is increased by groups. The last reshape layer restores the size of the output feature map to its original value.
    Because ABConv convolves a single kernel with the feature map that expands in the spatial direction, the weight arithmetic intensity is increased by $g$ times.
    
    Table~\ref{table:arithmetic_intensity} shows that the network capacity of ABConv may decrease since MACs and weight size of ABConv are reduced by the group number. In order to increase the network complexity of ABConv, we propose the expanded version of ABConv (ABConv-exp) which adds a pointwise convolution layer after the first reshape layer of ABConv. ABConv-exp sets the number of filters of pointwise convolution ($C_{mid}$) as $\frac{k^2C_{in}C_{out}}{C_{in}+k^2C_{out}}$ to have the same value as MACs of standard convolution.

    \subsection{Group selection algorithm}
	As shown in ABConv and ABConv-exp of Table~\ref{table:arithmetic_intensity}, as $g$ is increased excessively to increase the weight arithmetic intensity, the activation arithmetic intensity is rather decreased.
	Therefore, it is important to balance the weight data reuse and activation data reuse in order to increase overall arithmetic intensity.
    We propose a group selection algorithm to balance the weight and activation arithmetic intensity for increasing hardware performance and reducing latency (Algorithm~\ref{group_select}). 
    First, we calculate the quotients ($q_{in}, q_{out}$) and the remainders ($r_{in}, r_{out}$) of input and output channel directions when dividing channels by the step size ($t_{in}, t_{out}$) of the convolution staircase.
    Second, if the remainders are both zero, the optimal $g$ value ($g_{opt}$) that maximizes the overall arithmetic intensity is found by solving the problem of finding a balance point.
    When using ABConv and ABConv-exp, $is_{\text{exp}}$ is $False$ and $True$, respectively. 
    We list the common divisors of both quotients as group candidates. 
    Next, the value closest to $g_{opt}$ is selected among the group candidates.
    Lastly, the algorithm returns $g$ and $\text{sw}_{rep}$ which is a switch that decides whether or not to change the convolution to ABConv or ABConv-exp.

    \newcommand{\factorial}{\ensuremath{\mbox{\sc Factorial}}}
    \begin{algorithm}[t]
    \centering
    \caption{group selection algorithm}\label{group_select}
    \scriptsize
    \begin{algorithmic}[1]
    \Procedure{Group-select}{$S_o, k, C_{in},C_{out}, t_{in}, t_{out}, is_{\text{exp}}$}
        \State $q_{in} = floor(C_{in}/t_{in})$
        \State $q_{out} = floor(C_{out}/t_{out})$
        \State $r_{in} = C_{in}/t_{in} - q_{in}$
        \State $r_{out} = C_{out}/t_{out} - q_{out}$
        \If {$(r_{in}=0) \& (r_{out}=0)$}
            \If {$is_{\text{exp}} = False$}
                \State $g_{opt}$=$\sqrt{\frac{k^2C_{in}C_{out}}{S_o^2(C_{in}+C_{out})}}$
            \Else
                \State $g_{opt}$=$\frac{-C_{in}-C_{out}+\sqrt{\frac{{(C_{in}+C_{out})^2+8(C_{in}+k^2C_{out})C_{mid}^2}}{S_o^2}}}{4C_{mid}}$
            \EndIf
            \State $G$ = Common divisor$(q_{in},q_{out})$
            \State $g=G[argmin(|G-g_{opt}|)]$
            \If {$g=1$}
                \State $\text{sw}_{rep}=False$
            \Else
                \State $\text{sw}_{rep}=True$
            \EndIf
        \Else
            \State $g = 1$, $\text{sw}_{rep} = False$
        \EndIf
    \State \Return $g$, $\text{sw}_{rep}$
    \EndProcedure
    \end{algorithmic}
    \end{algorithm}

\section{Experimental Setup}
    \subsection{Target devices}
    The experiments focused on edge devices and used Arm Ethos-U65 Micro-NPU \cite{U65} as the NPU device.
    We made use of Arm Fixed Virtual Platforms (FVP) \cite{FVP} to conduct a complete simulation of an Arm NPU-integrated system.
    Ethos-U65 device is configured with 256 MAC processing units and dedicated SRAM memory mode for our experiments.
    The models for Ethos-U65 device were quantized into 8-bit integer and converted through Vela compiler \cite{Vela}.
    
    \subsection{Neural networks}
    As bottleneck layers in efficient models for edge devices are typically composed of pointwise convolutions $(k=1)$, we first validated our proposed method by replacing them in this study.
    We chose MobileNetV1 and ResNet50 as the baseline model and applied the group selection algorithm to each pointwise convolution layer of the baseline model. If the algorithm returns $\text{sw}_{rep}$ as $True$ and $g$ is greater than 1, we replaced the pointwise convolution layer with ABConv or ABConv-exp.
    The experimental evaluation is performed on CIFAR100 dataset to focus on the small spatial size ($S_o = 32$). 
    
    We use the pre-trained baseline models and train the variants model for 200 epochs using batches of size 128.
    We use the standard stochastic gradient descent optimizer with momentum set to 0.9. 
    We use an initial learning rate of 0.001, and reduce the learning rate when the accuracy of the validation set has stopped improving within 0.0001.
    We measure the latency of inference when the batch size is 1.

\section{Results}
    \subsection{Hardware performance}
    \begin{figure}[t]%
    \centering
        \includegraphics[width=0.48\columnwidth ]{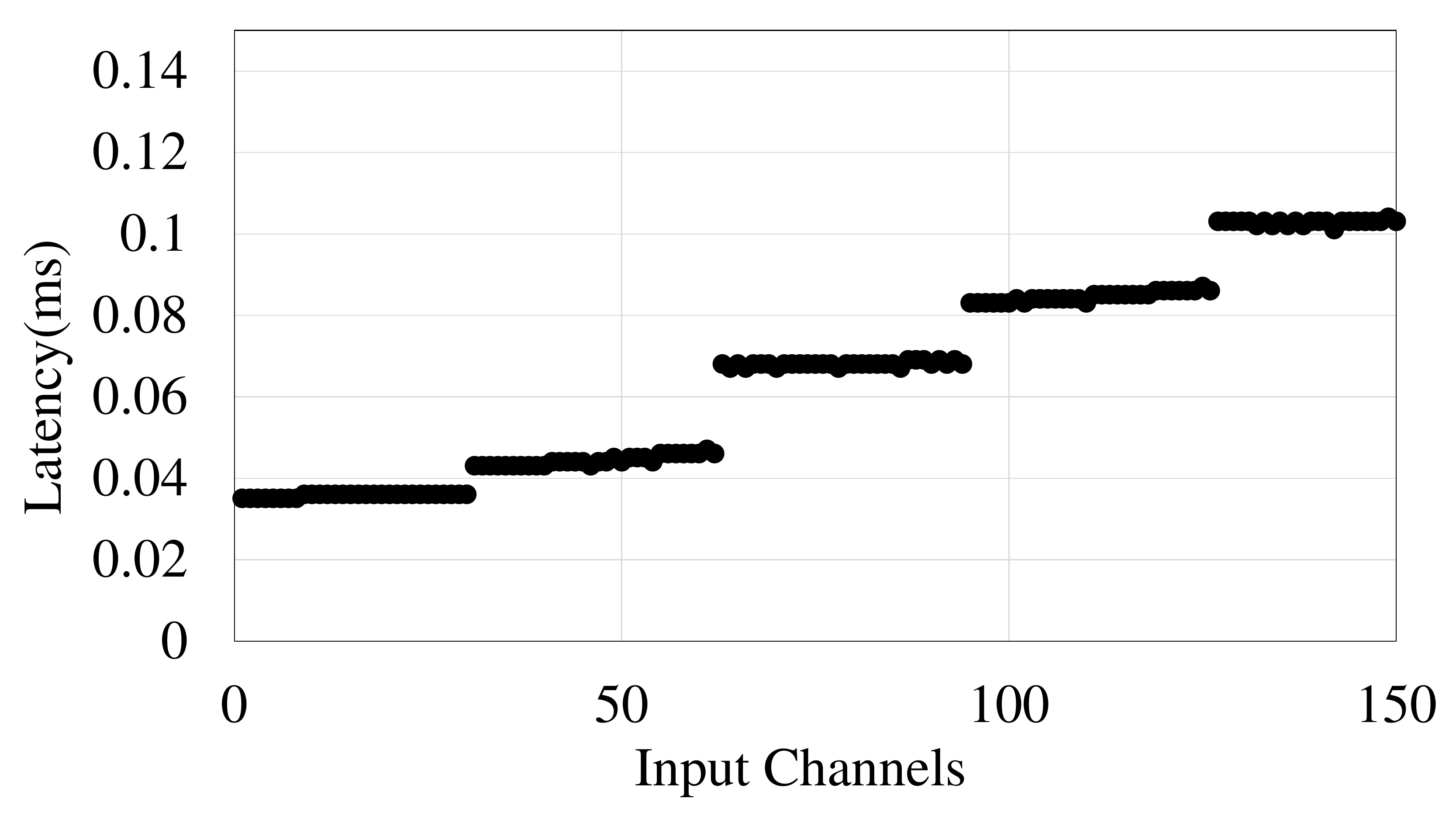} 
        \includegraphics[width=0.48\columnwidth ]{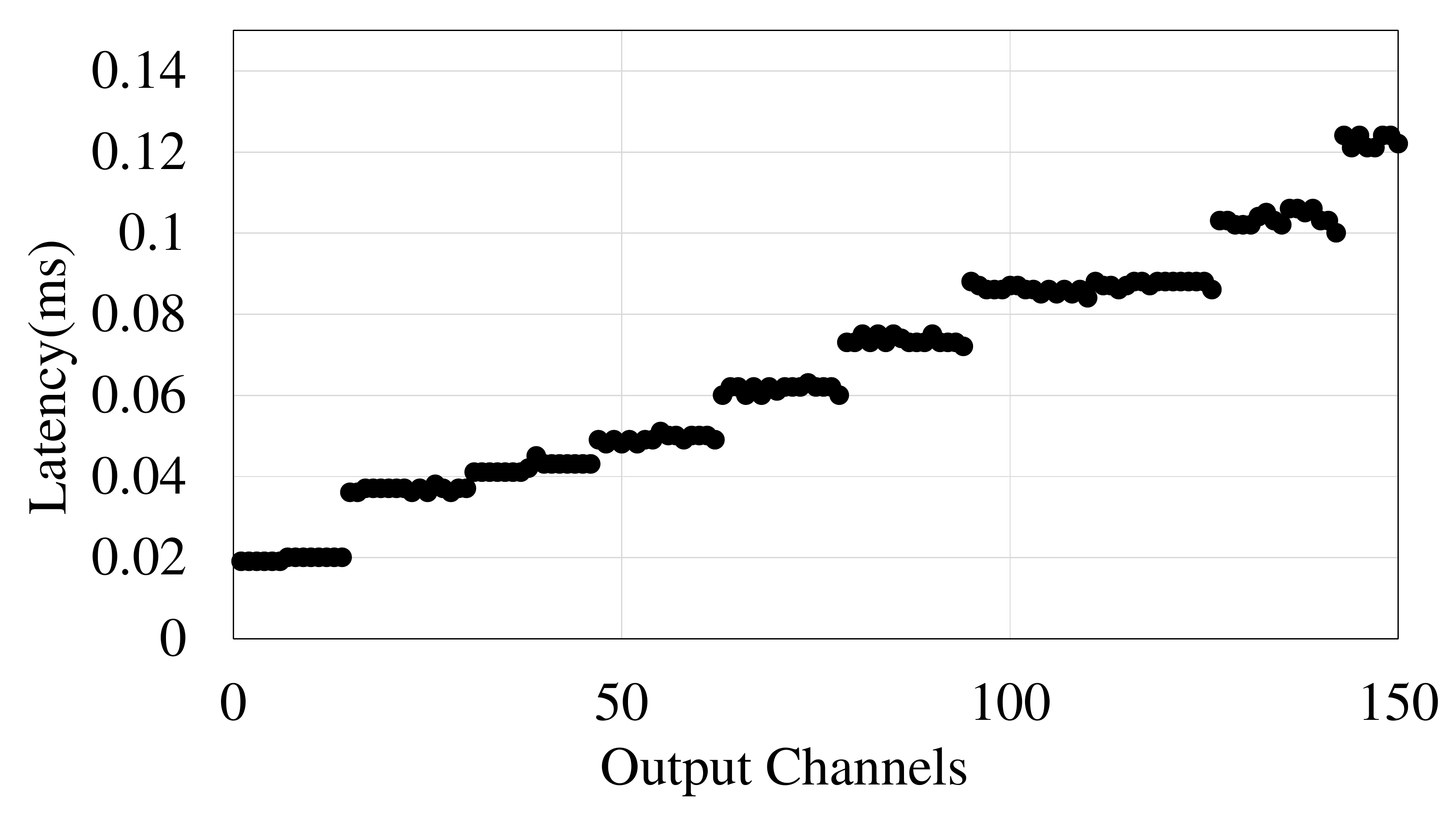} 
        \caption{Staircase latency plots when ${S_o}$ is 32.}%
    \label{Staircse}%
    \end{figure}

    In order to set our grouping selection algorithm properly, the method should aware step sizes of devices for each convolution parameters. 
    As shown in Figure~\ref{Staircse}, convolution operation on Ethos-U65 have step size of 32 in the input channels ($t_{in}$) and 16 in the output channels ($t_{out}$). 
     
    \begin{figure}[t]%
        \includegraphics[width=0.48\columnwidth ]{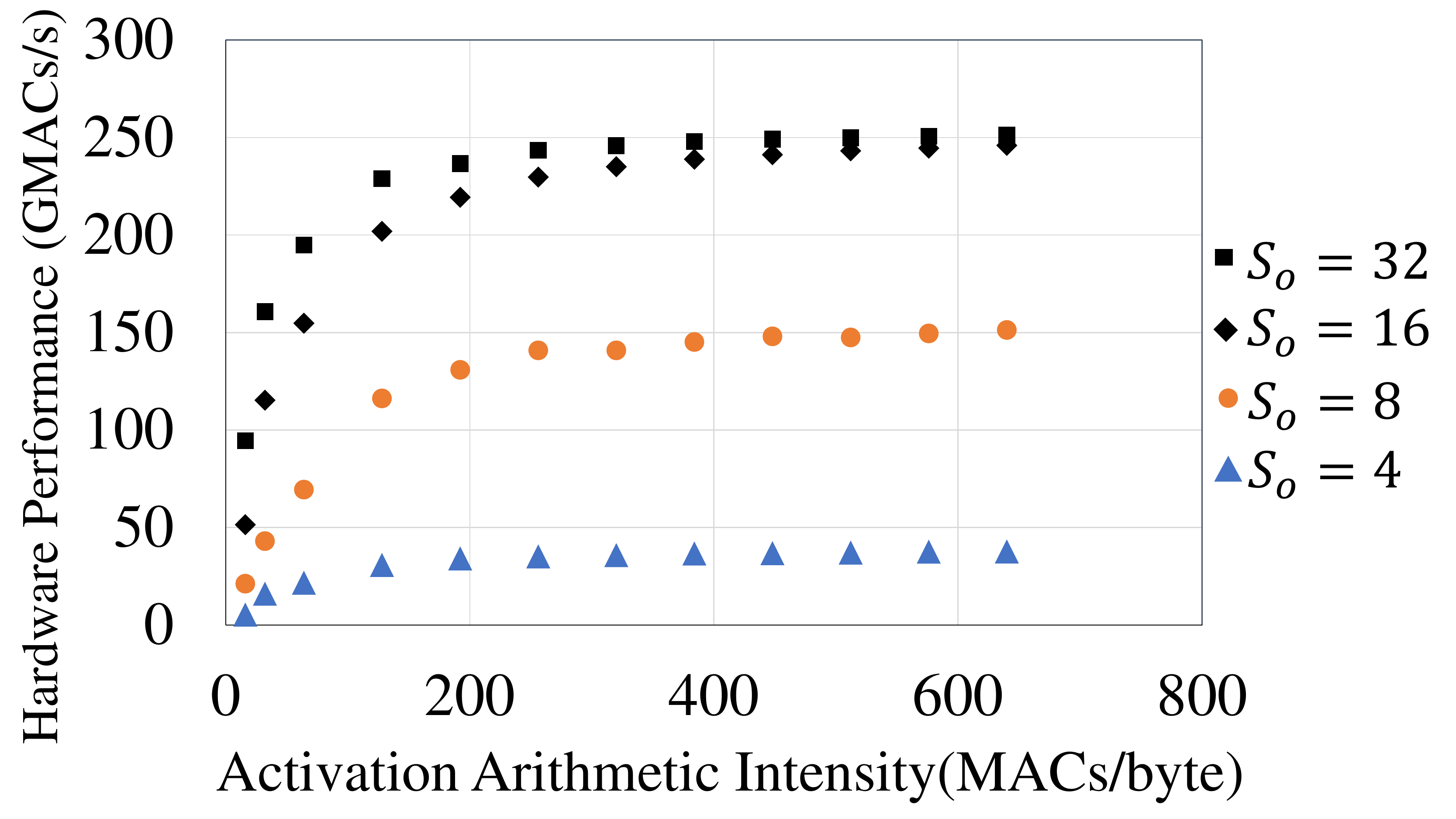} 
        \includegraphics[width=0.48\columnwidth ]{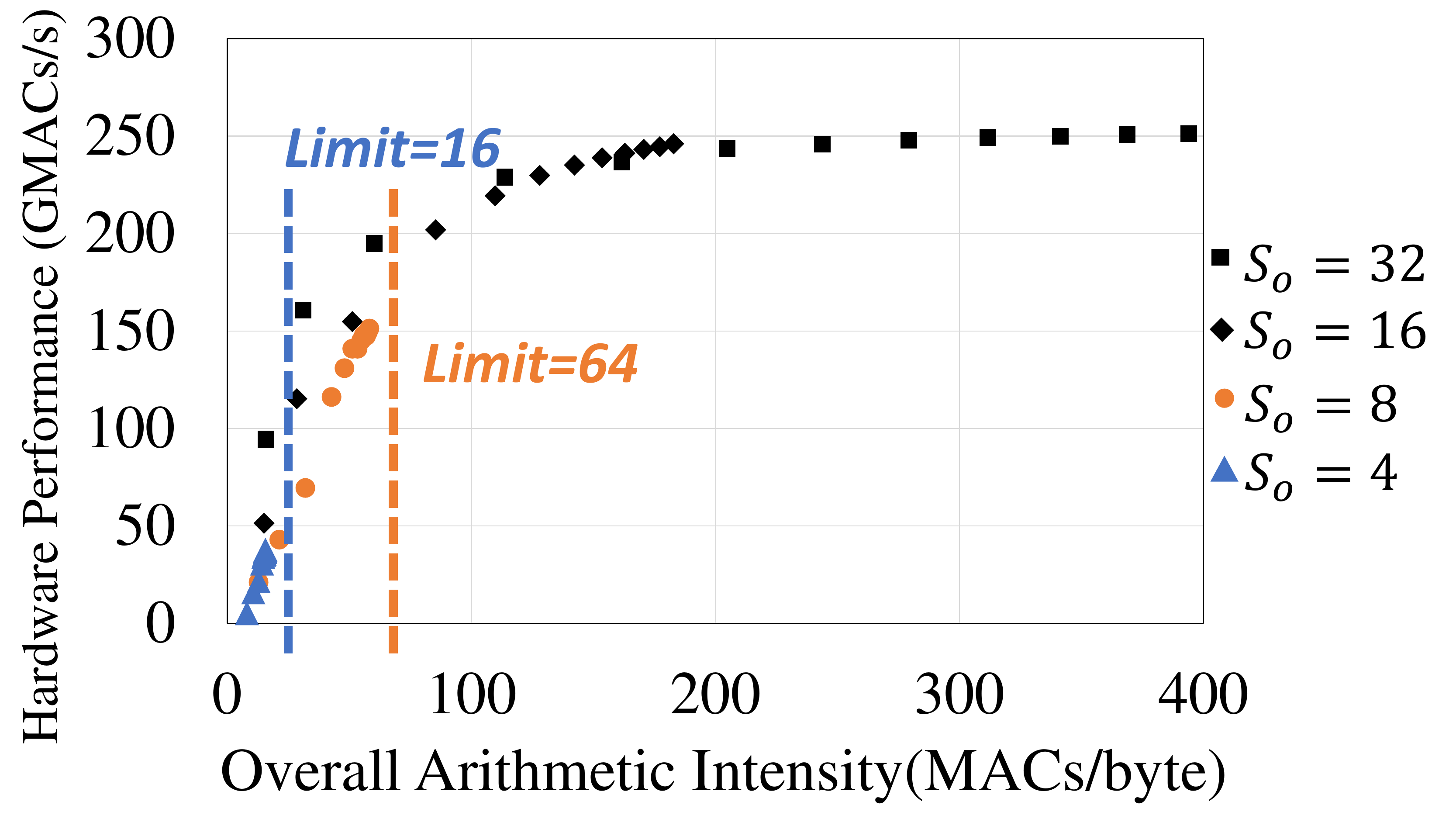} 
        \caption{Roofline analysis for pointwise convolutions. Y-axis presents the Giga-MACs per second.}%
    \label{Roofline_pointwise}%
    \end{figure}        

    Before the evaluation of ABConv, we investigated on the inefficiency of pointwise convolutions, especially for small spatial sizes.
    Figure~\ref{Roofline_pointwise} plots the roofline figures of Ethos-U65 with respect to arithmetic intensity of the activation data and overall arithmetic intensity. 
    Each point represent one pointwise convolution, varying their in/out channels and spatial sizes.
    We observed that when the spatial size is small, such as 4 and 8, the hardware performance of the convolution is bounded to a certain limit.        
    This is because the overall arithmetic intensity is bounded to the arithmetic intensity of weight data, even if they have enough intensity on the activation data.
    The plots with small spatial size cannot go over certain limits, which are the square of the spatial size.
    Therefore, no matter how well the convolution reuse the activation data, they cannot overcome the upper limit of overall arithmetic intensity, resulting in low hardware performance as long as they have a small spatial size.

    \begin{table}[t]
    \centering
    \caption{Comparison of pointwise convolution and proposed method when the configuration of $S_o^2\times C_{in}\times C_{out}$ is $4^2\times1024\times1024$}
    \label{table:arithmetic_intensity_and_latency}
    \resizebox{\linewidth}{!}{%
    \begin{tabular}{c|c|c|r|r|c|c|c|r} 
    \toprule
    \multirow{2}{*}{}           & \multirow{2}{*}{\begin{tabular}[c]{@{}c@{}}group\\policy\end{tabular}} & \multirow{2}{*}{$g$} & 
    \multirow{2}{*}{\begin{tabular}[c]{@{}c}MACs\end{tabular}}
    & 
        \multirow{2}{*}{\begin{tabular}[c]{c}Params\end{tabular}}

    & \multicolumn{3}{c|}{Arithmetic intensity} &  \multirow{2}{*}{\begin{tabular}[c]{@{}c@{}}Latency\\($\mu$s)\end{tabular}}   \\ 
    \cline{6-8}
                                &                                                                        &                         &                       &                         & Weight & Activation & Whole       &                                         \\ 
    \toprule
    Pointwise Conv               &                                                                        &                         & 16,777,216          & 1,048,576             & 16     & 512        & 15.5                &  452.1  \\                             
    \hline
    \multirow{2}{*}{ABConv}     &                                                                        & 32                      & 524,288          &  1,024            & 512    & 16         & 15.5                & 7.1 \\                              
    \cline{2-9}
                                & $\checkmark$                                                                      & 4                       &  4,194,304          &  65,536             & 64     & 128        & 42.7                &  36.1  \\                              
    \hline
    \multirow{2}{*}{ABConv-exp} &                                                                        & 32                      & 16,777,216          &  32,768             & 512    & 30         & 28.4                &  110.1  \\                               
    \cline{2-9}
                                & $\checkmark$                                                                      & 8                       & 16,777,216          &  131,072             & 128    & 102        & 56.9                &  86.1  \\                             
    \bottomrule
    \end{tabular}
    }
    \end{table}

    \subsection{ABConv}
    Table~\ref{table:arithmetic_intensity_and_latency} reports the comparison of the convolution layer, the grouped convolution, ABConv and ABConv-exp with and without the group selection algorithms. ABConv and ABConv-exp have larger arithmetic intensity and lower latency than the convolution layer and the group convolution by increasing the weight arithmetic intensity. Without the group selection algorithm, the arithmetic intensity of weight data is increased with the largest $g$ as 32, the whole arithmetic is decreased by bounding to the reduced activation arithmetic intensity due to the largest $g$. 
    In the ABConv without a group selection algorithm, it has the lowest latency, but the reason is that MACs decreased due to large $g$ values.

    \begin{figure}[t]
    \centering
    \begin{subfigure}[b]{0.48\columnwidth}
             \centering
             \includegraphics[width=\columnwidth]{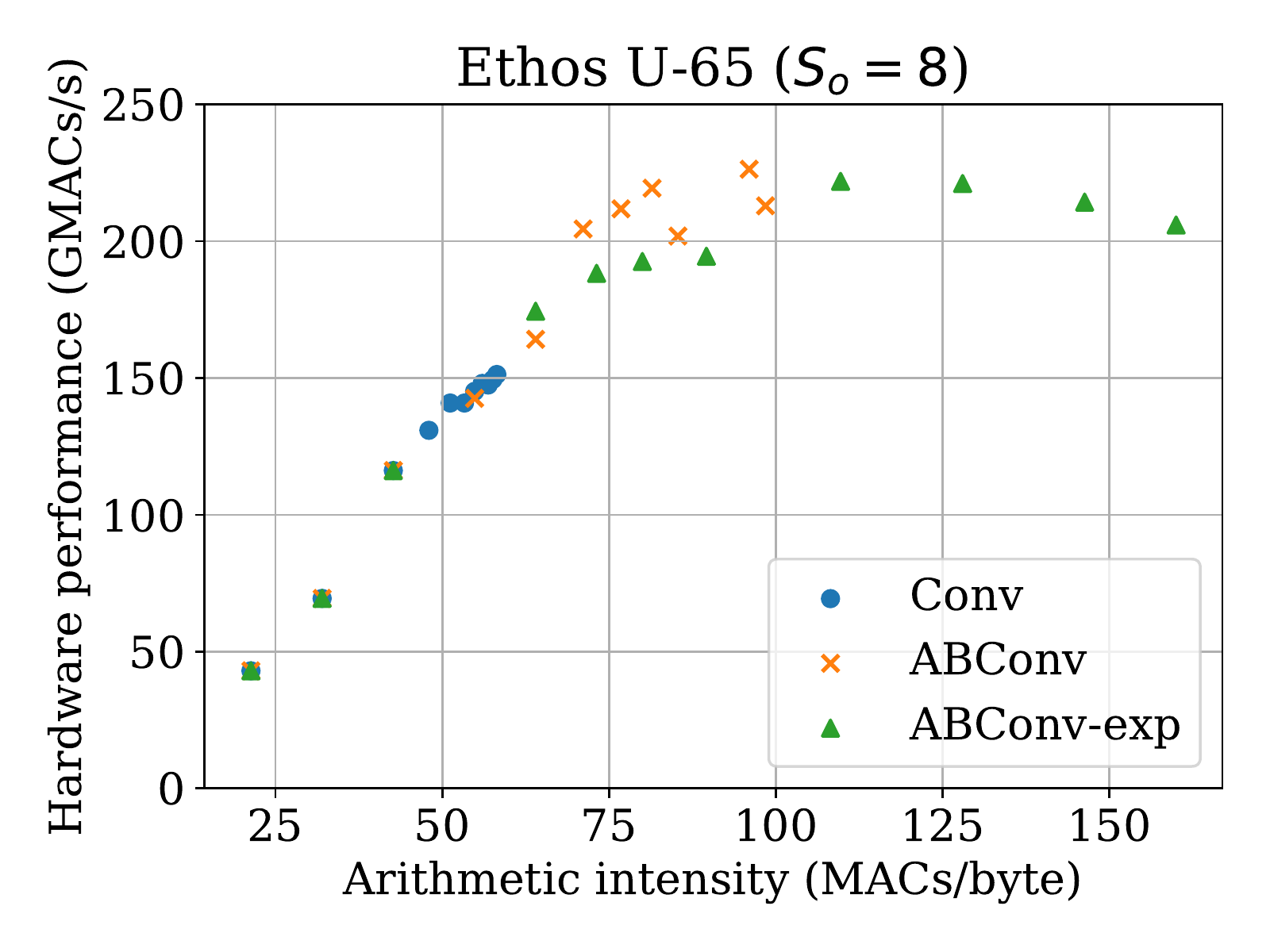}
         \end{subfigure}
    \begin{subfigure}[b]{0.48\columnwidth}
             \centering
             \includegraphics[width=\columnwidth]{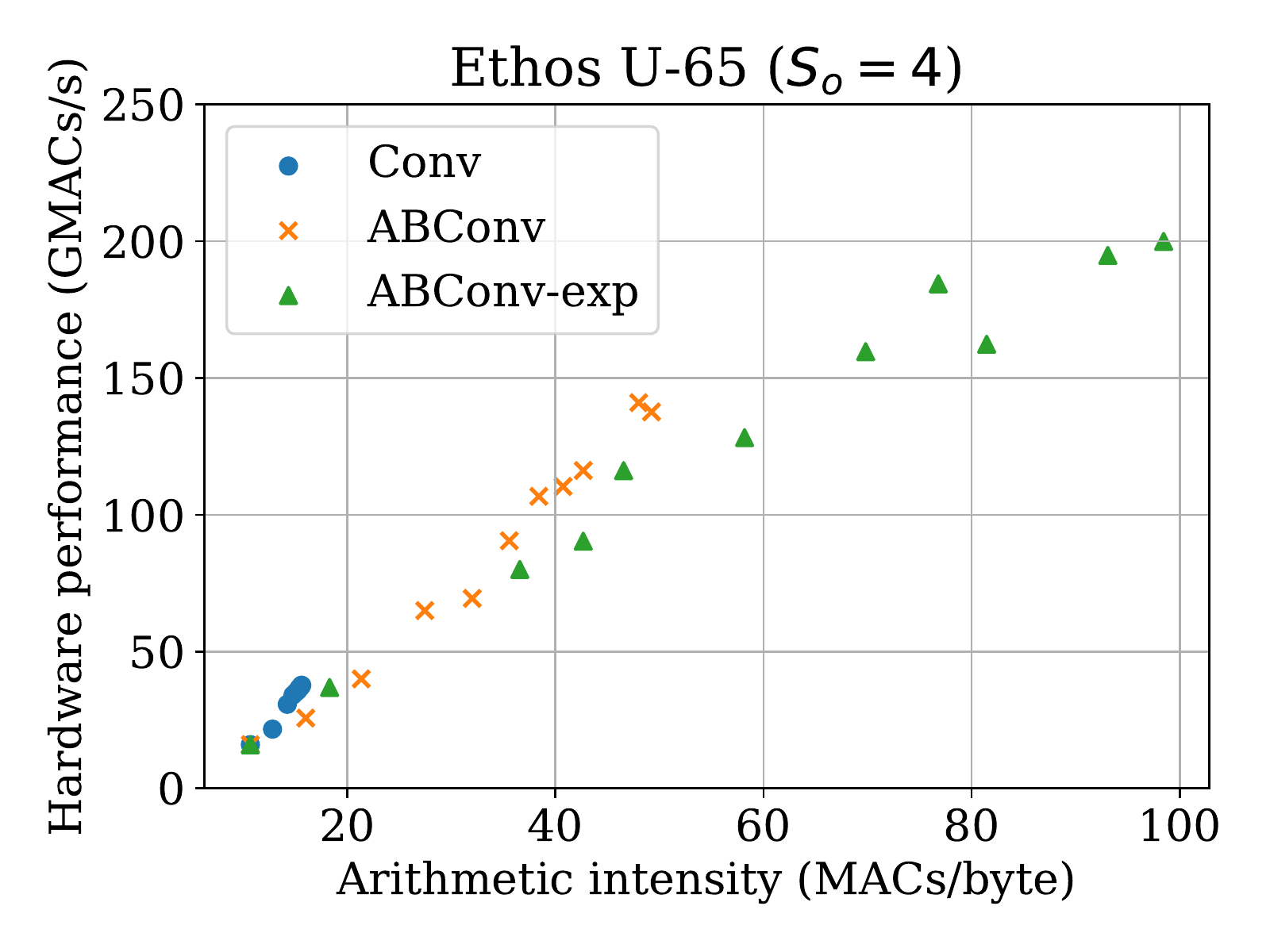}
         \end{subfigure}
    \caption{Roofline analysis on Ethos-U65 when $S_o$ is 4 and 8. Y-axis presents the Giga-MACs per second.}
    \label{Fig:roofline_ABConv}
    \end{figure}
x
    Figure~\ref{Fig:roofline_ABConv} shows the roof line analysis of ABConv and ABConv-exp on Ethos-U65 when the channel input and channel output have a configuration that increases by 128 from 128 to 1280. It also report that the arithmetic intensity of ABConv and ABConv-exp is larger than that of the convolution layer. Therefore, the hardware performance of both proposed method have larger values than that of the convolution layer. 
    As the overall arithmetic intensity decreases more with smaller input sizes, the proposed convolution layers show greater improvement in hardware performance.

\begin{figure}[t]
    \centering
    \begin{subfigure}[b]{0.48\columnwidth}
             \centering
             \includegraphics[width=\columnwidth]{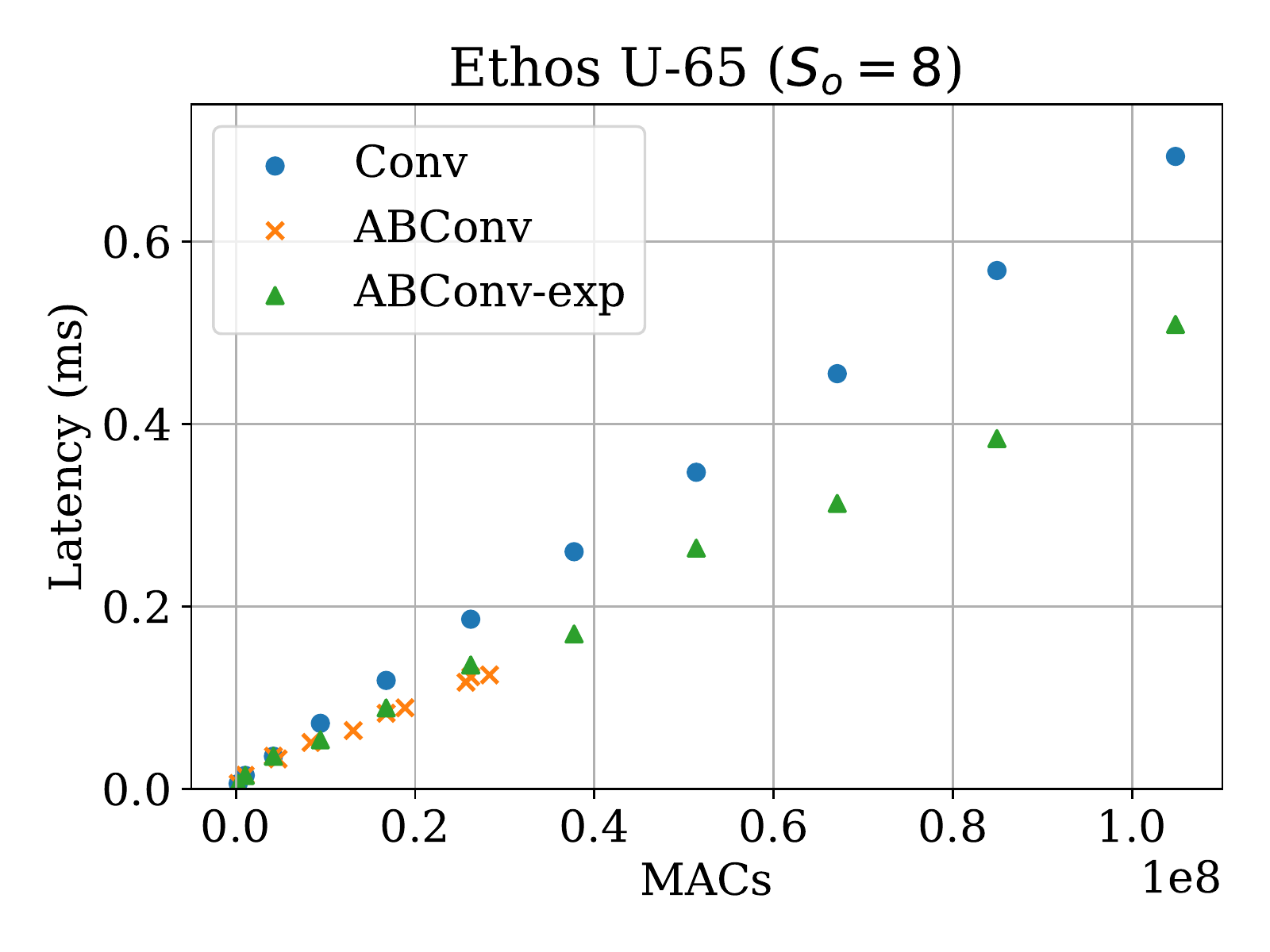}
         \end{subfigure}
    \begin{subfigure}[b]{0.48\columnwidth}
             \centering
             \includegraphics[width=\columnwidth]{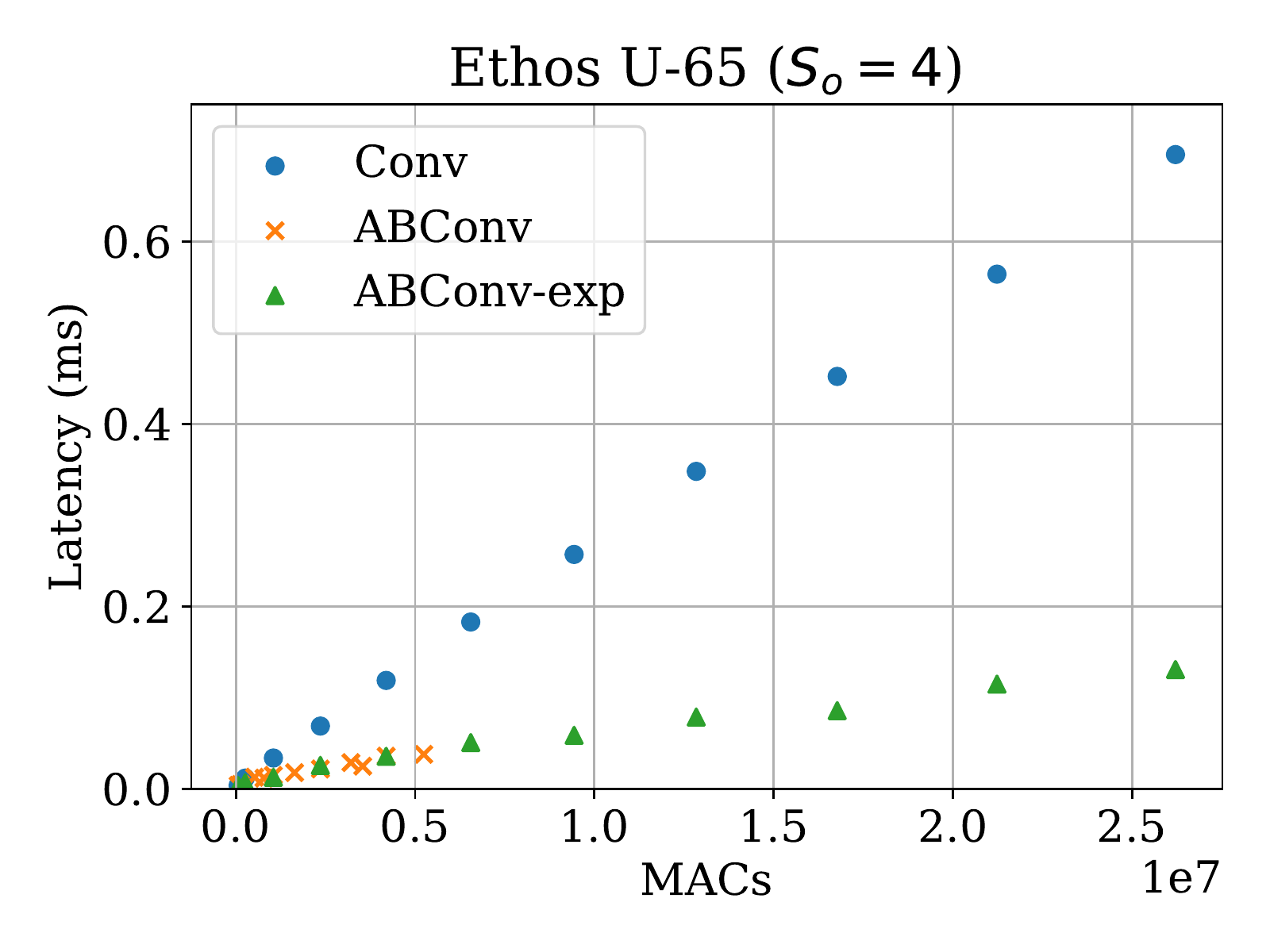}
         \end{subfigure}
    \caption{Latency vs MACs on Ethos-U65 when $S_o$ is 4 and 8.}
    \label{Fig:latency_vs_macs_ABConv}
    \end{figure}

    Figure~\ref{Fig:latency_vs_macs_ABConv} presents the latency on Ethos-U65 versus the number of MACs when the configuration is set to the same as Figure~\ref{Fig:roofline_ABConv}. 
    Since the proposed ABConv and ABConv-exp have high performance in terms of hardware utilization, they have low latency. 
    Compared to the convolution layer, ABConv has low MACs and latency, and ABConv-exp has the same MAC but lower latency.
        
    To involve the efficient layer in the baseline networks, we replace the existing pointwise convolution layers with ABConv to improve the hardware performance. 
    However, replacing all pointwise convolutions with ABConv decreases the accuracy due to a lack of communication between different groups of channels. Therefore, it is necessary to alternate between pointwise convolutions and ABconv to maintain accuracy.
    
    As shown in Table~\ref{table:final_acc}, we compare the performance of two baseline models by replacing each pointwise convolution with ABConv and ABConv-exp. Based on MobileNetV1, the models with the proposed layers increase accuracy and decrease latency.
    We observe that the modified models in ResNet50 also reduce the latency with a tiny drop in accuracy.

\begin{table}[t]
\centering
\caption{CIFAR100 performance comparison of the models}
\label{table:final_acc}
\resizebox{.9\linewidth}{!}{%
\begin{tabular}{c|c|c|r|r|r} 
\toprule
Baseline & Type & \begin{tabular}[c]{c}Accuracy\\(\%)\end{tabular} & \begin{tabular}[c]{@{}c@{}}MACs\\(M)\end{tabular} & \begin{tabular}[c]{@{}c@{}}Params\\(M)\end{tabular} & \begin{tabular}[c]{@{}c@{}}Latency\\(ms)\end{tabular} 
\\
\toprule
\multirow{3}{*}{MobileNetV1} & original              & 66.73                                                                   & 46.5                                                               & 3.3                                                                  & 1.43                      \\ 
\cline{2-6}
                             & ABConv                & 67.18                                                                   & 39.9                                                               & 2.4                                                                  & 0.99                      \\ 
\cline{2-6}
                             & ABConv-exp            & 66.95                                                                   & 46.8                                                               & 2.4                                                                  & 1.02                      \\ 
\hline
\multirow{3}{*}{ResNet50}    & original              & 78.94                                                                   & 1298.0                                                             & 23.8                                                                 & 12.35                     \\ 
\cline{2-6}
                             & ABConv                & 78.73                                                                   & 1239.3                                                             & 21.0                                                                 & 11.56                     \\ 
\cline{2-6}
                             & ABConv-exp            & 78.69                                                                   & 1304.8                                                             & 21.5                                                                 & 11.91                      \\
\bottomrule
\end{tabular}}
\end{table}

\section{Conclusions}
In this paper, we report that the overall arithmetic
intensity is bound to the weight arithmetic intensity as weight reuse is reduced. We propose ABConv as a new layer that balances the arithmetic intensity of weight data and activation data, maximizing the whole arithmetic intensity.
To validate ABConv, we replaced the pointwise convolutions in MobileNetV1 and ResNet50 with ABConv.
We observe that the models using ABConv reduce the latency on Arm Ethos-U65 NPU without an accuracy drop on CIFAR100. 
Future research can compare experiments between the standard convolution ($k>1$) and the proposed method to further validate the proposed method.


\bibliography{main}
\bibliographystyle{mlsys2023}

\clearpage
\appendix
\begin{large}
\textbf{\textsc{Appendix}}
\end{large}
\section{Ablation study}

Table~\ref{table:architecture_study} shows the trade-off between model accuracy on CIFAR100 and latency of the different models mixing the pointwise convolution and the proposed method. $P$, $A$, and $E$ denote the model using the pointwise convolution, ABConv, and ABConv-exp alternately. As we expected, the model using the ABConv only shows the fastest latency, but it reduces the accuracy due to the lack of communication between different groups of channels. We observed that the models mixing $P$, $A$, and $E$ connect between channels and increase the accuracy.

\section{NVIDIA Jetson Nano}
\begin{figure}[ht]%
\centering
    \includegraphics[width=0.48\columnwidth ]{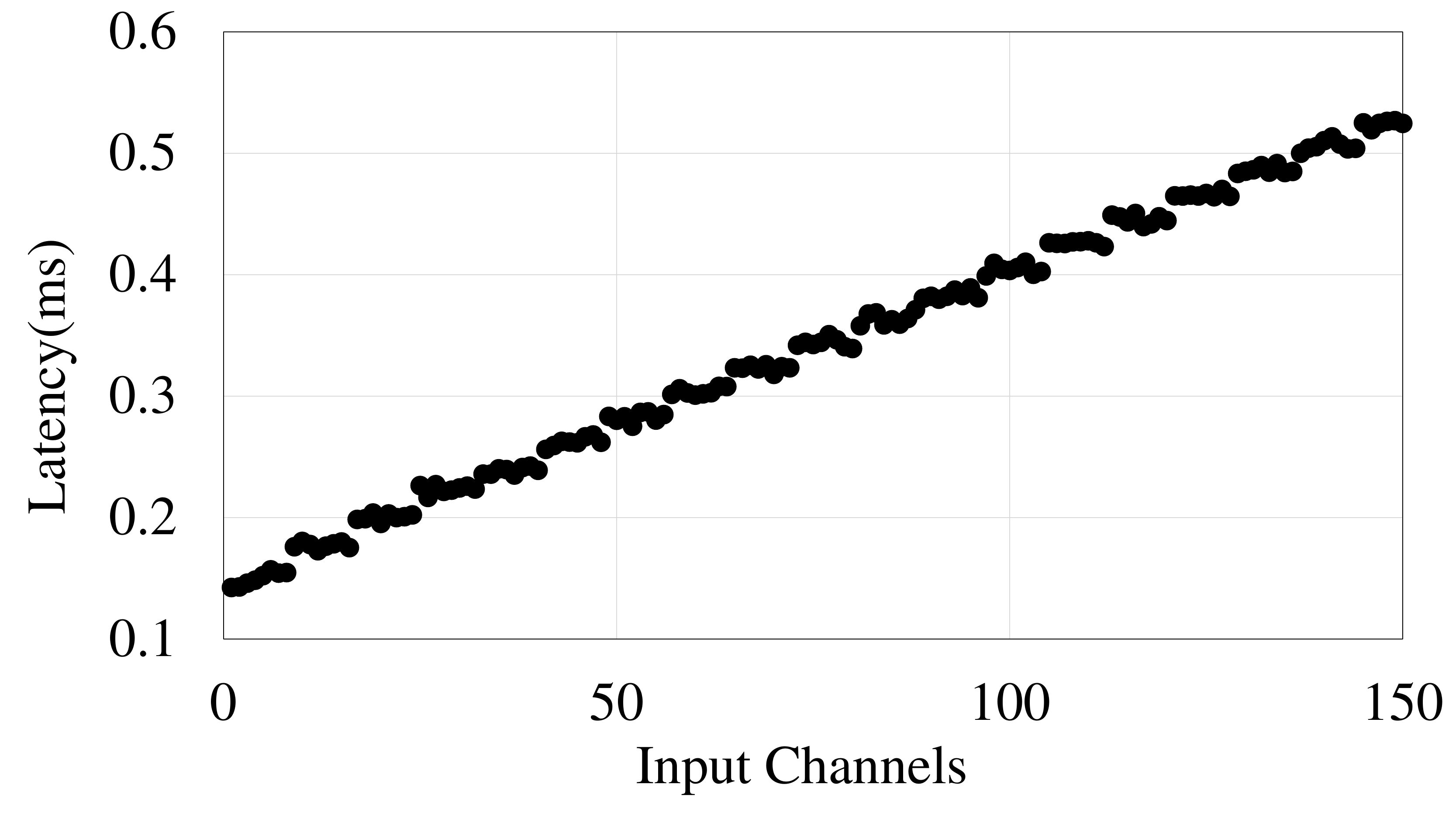} 
    \includegraphics[width=0.48\columnwidth ]{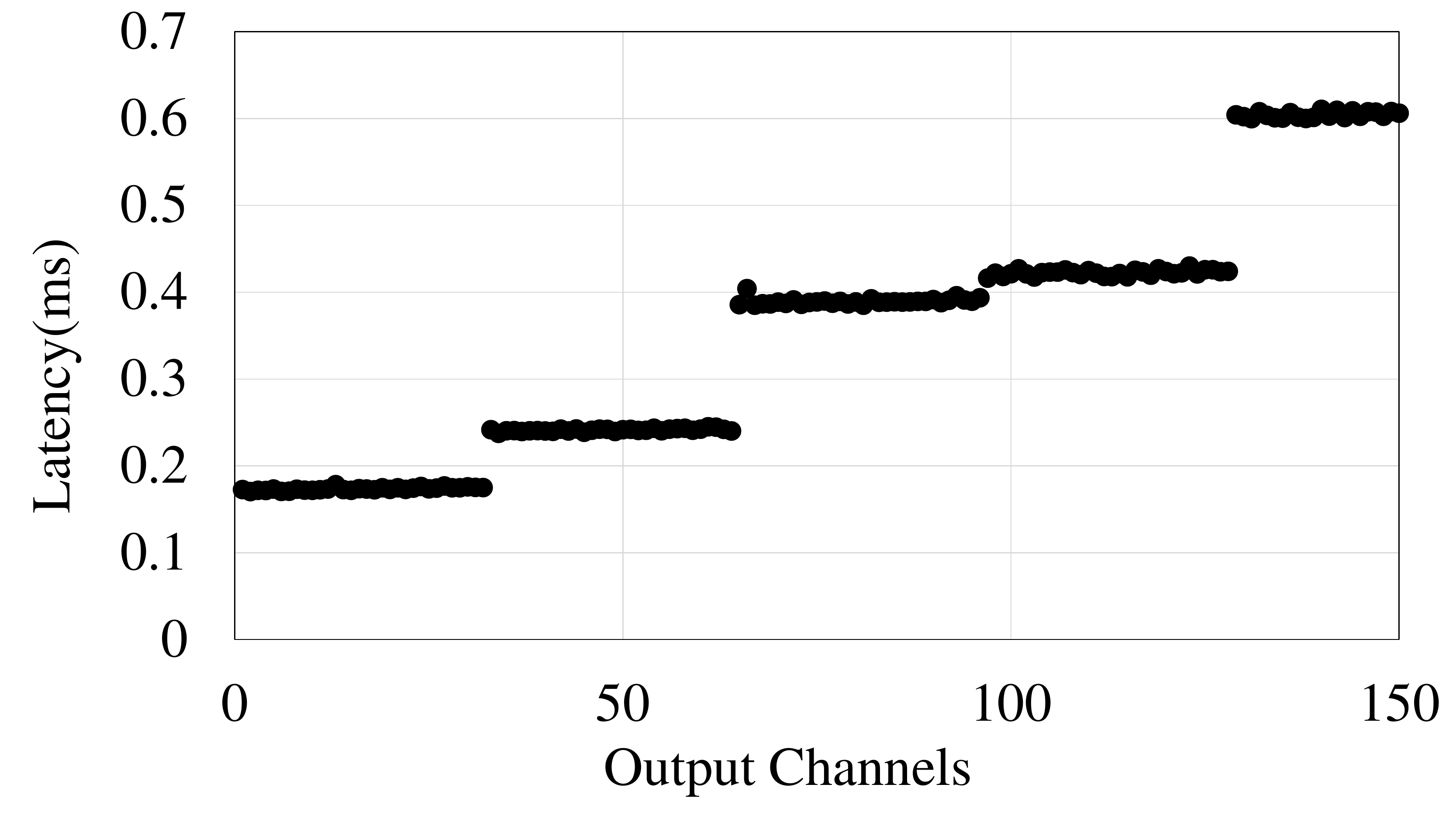} 
    \caption{Staircase latency plots on NVIDIA Jetson Nano when ${S_o}$ is 32.}%
\label{Staircse_nano}%
\end{figure}

\begin{figure}[ht]
\centering
\begin{subfigure}[b]{0.48\columnwidth}
         \centering
         \includegraphics[width=\columnwidth]{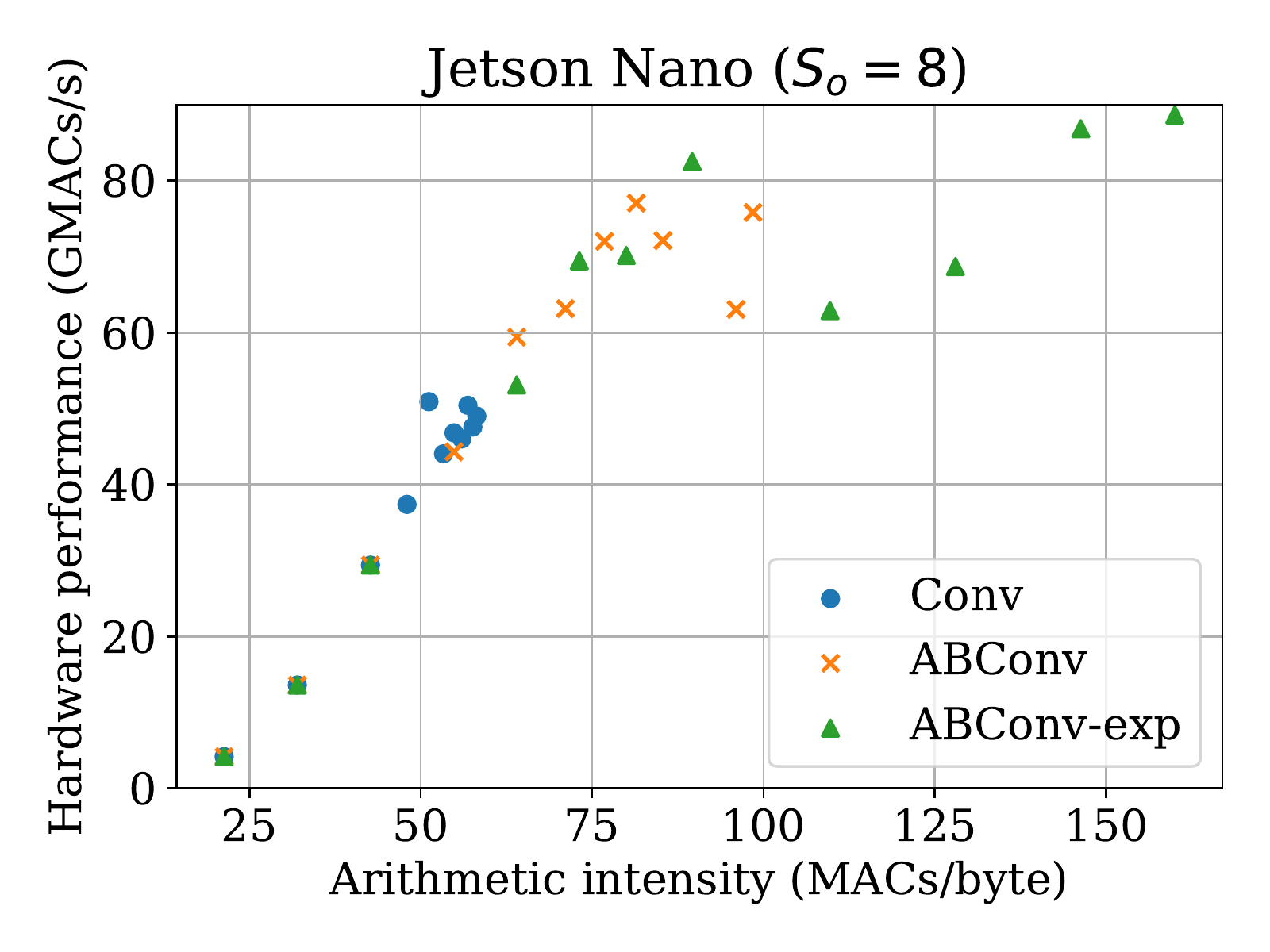}
     \end{subfigure}
\begin{subfigure}[b]{0.48\columnwidth}
         \centering
         \includegraphics[width=\columnwidth]{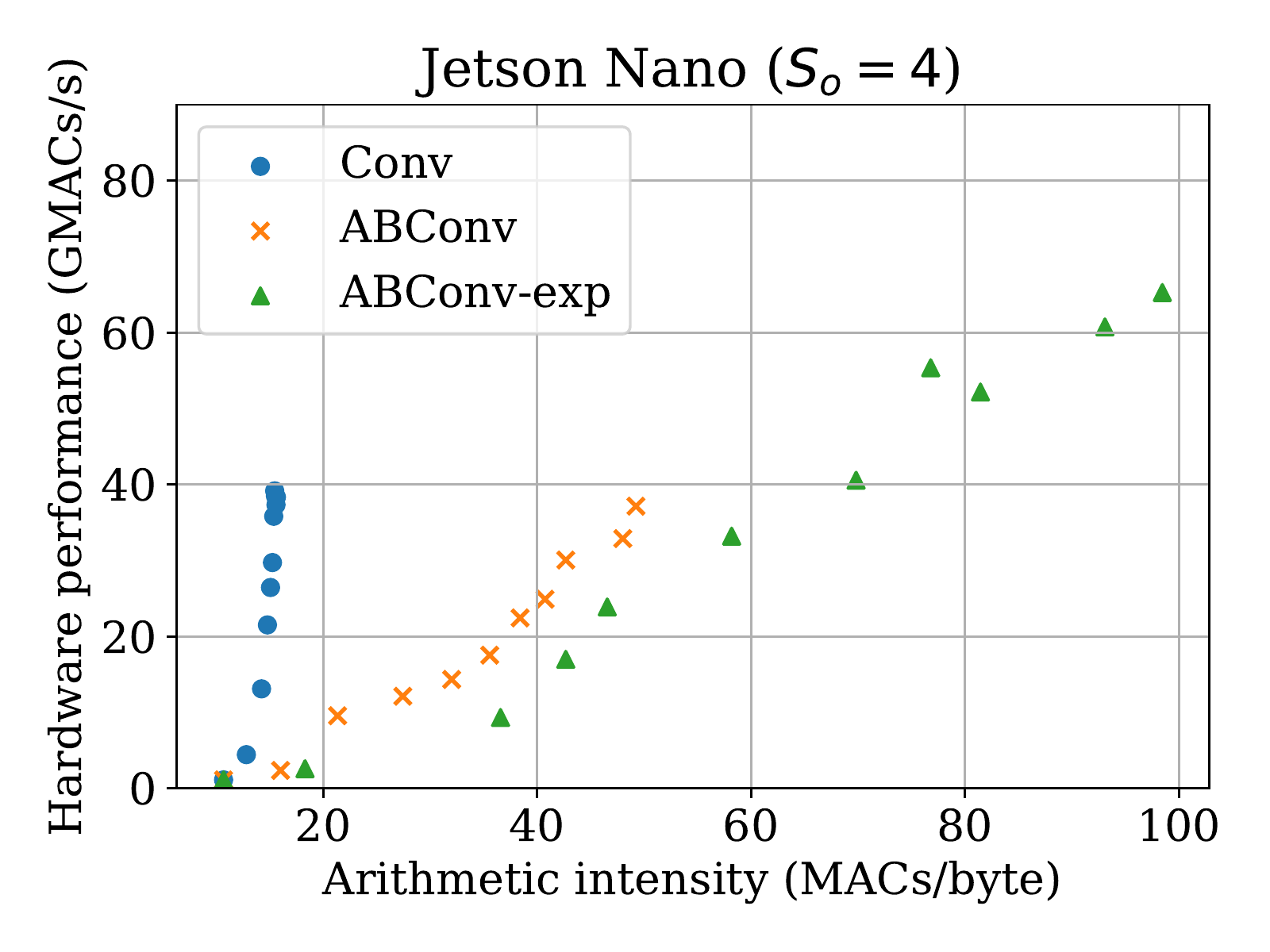}
     \end{subfigure}
\caption{Roofline analysis on Jetson Nano when $S_o$ is 4 and 8. Y-axis presents the Giga-MACs per second.}
\label{Fig:roofline_ABConv_nano}
\end{figure}

\begin{figure}[ht]
\centering
\begin{subfigure}[b]{0.48\columnwidth}
         \centering
         \includegraphics[width=\columnwidth]{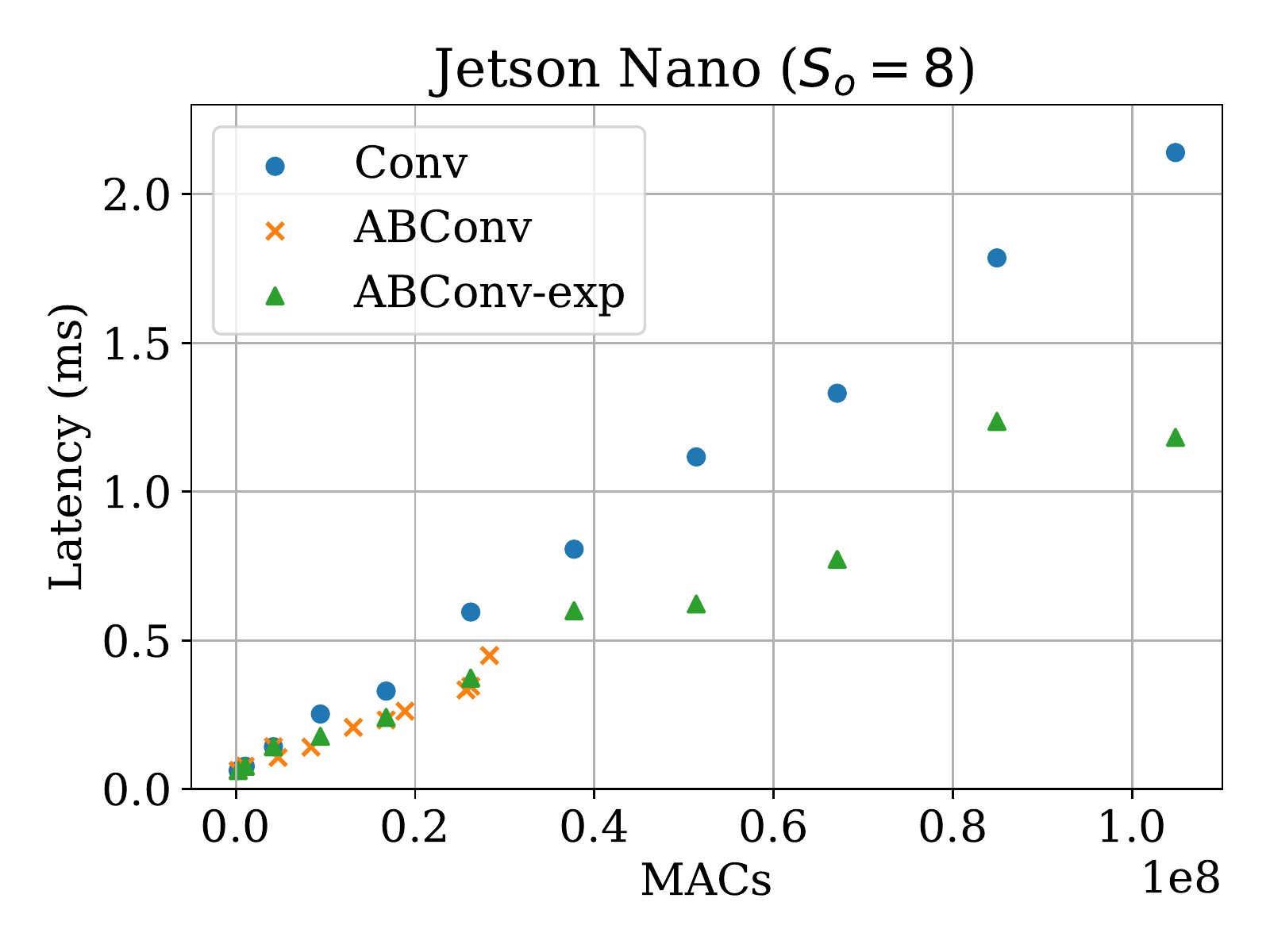}
     \end{subfigure}
\begin{subfigure}[b]{0.48\columnwidth}
         \centering
         \includegraphics[width=\columnwidth]{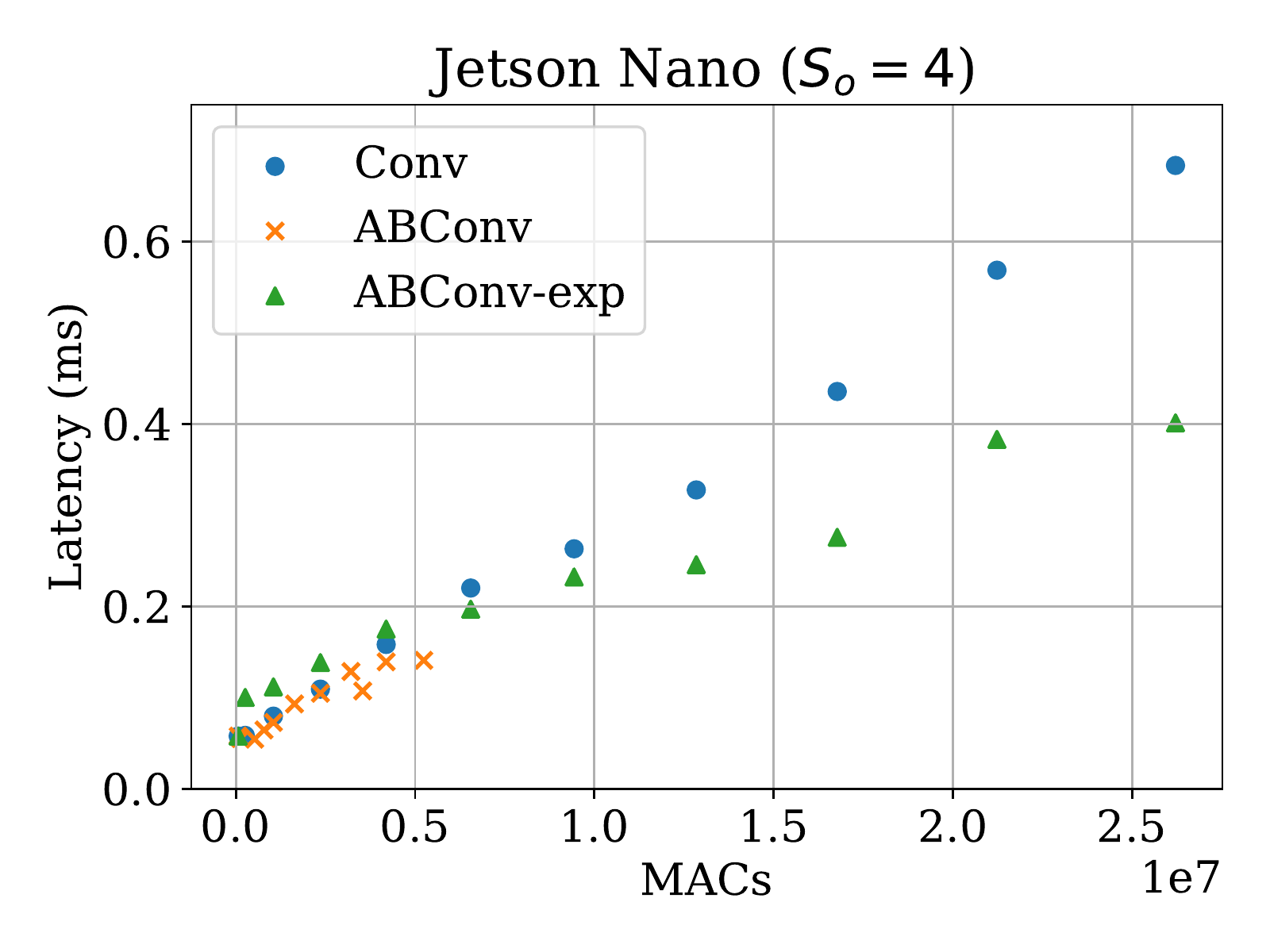}
     \end{subfigure}
\caption{Latency vs MACs on Jetson Nano when $S_o$ is 4 and 8.}
\label{Fig:latency_vs_macs_ABConv_nano}
\end{figure}

To validate the proposed ABConv on different devices, we chose the NVIDIA Jetson Nano as an edge GPU device and conducted the same experiments as with Ethos U-65 NPU. 
As shown in Figure~\ref{Staircse_nano}, convolution operations on Jetson Nano have step size of 8 in the input channels and 32 in the output channels. 
Figure~\ref{Fig:roofline_ABConv_nano} and \ref{Fig:latency_vs_macs_ABConv_nano} demonstrate that the ABConv and ABConv-exp also have higher arithmetic intensity and reduce latency with the same MACs on Jetson Nano.
Table~\ref{table:final_acc_nano} shows that MobileNetV1 and ResNet50 by replacing each pointwise convolution with ABConv and ABConv-exp reduce the latency with a similar accuracy.

\begin{table}[hb]
\centering
\caption{Architecture study of MobileNetV1 on Ethos-U65}
\label{table:architecture_study}
\resizebox{.9\columnwidth}{!}{%
\begin{tabular}{l|c|c|c} 
\toprule
\multicolumn{1}{c|}{Models} & MACs (M) & Latency (ms) & Accuracy (\%)  \\ 
\toprule
$P-P-P-P-P-P$               & 46.5     & 1.43         & 66.73          \\ 
\hline
$A-A-A-A-A-A$               & 23.6     & 0.42         & 64.72          \\ 
\hline
$A-P-A-P-A-P$               & 36.8     & 0.95         & 66.69          \\ 
\hline
$A-P-P-A-P-P$               & 39.9     & 0.99         & 67.18          \\ 
\hline
$E-E-E-E-E-E$               & 46.8     & 0.60         & 65.26          \\ 
\hline
$E-P-E-P-E-P$               & 46.8     & 1.02         & 66.95          \\ 
\hline
$E-P-P-E-P-P$               & 46.8     & 1.07         & 66.92          \\
\bottomrule
\end{tabular}}
\end{table}

\begin{table}[hb]
\centering
\caption{CIFAR100 performance comparison of the models on Jetson Nano}
\label{table:final_acc_nano}
\resizebox{.9\linewidth}{!}{%
\begin{tabular}{c|c|c|r|r|r} 
\toprule
Baseline & Type & \begin{tabular}[c]{@{}c@{}}Accuracy\\(\%)\end{tabular} & \begin{tabular}[c]{@{}c@{}}MACs\\(M)\end{tabular} & \begin{tabular}[c]{@{}c@{}}Params\\(M)\end{tabular} & \begin{tabular}[c]{@{}c@{}}Latency\\(ms)\end{tabular}
\\
\toprule
\multirow{3}{*}{MobileNetV1} & original              & 66.73                                                                   & 46.5                                                               & 3.3                                                                  & 2.94                      \\ 
\cline{2-6}
                             & ABConv                & 67.18                                                                   & 39.9                                                               & 2.4                                                                  & 2.79                      \\ 
\cline{2-6}
                             & ABConv-exp            & 66.95                                                                   & 46.8                                                               & 2.4                                                                  & 2.94                      \\ 
\hline
\multirow{3}{*}{ResNet50}    & original              & 78.94                                                                   & 1298.0                                                             & 23.8                                                                 & 23.09                     \\ 
\cline{2-6}
                             & ABConv                & 78.73                                                                   & 1239.3                                                             & 21.0                                                                 & 20.97                     \\ 
\cline{2-6}
                             & ABConv-exp            & 78.69                                                                   & 1304.8                                                             & 21.5                                                                 & 21.69                      \\
\bottomrule
\end{tabular}}
\end{table}

\end{document}